\documentclass[runningheads]{llncs}
\usepackage[T1]{fontenc}
\usepackage{url}
\usepackage[hidelinks]{hyperref} 

\usepackage{graphicx}
\usepackage{amsmath}
\usepackage{orcidlink}
\begin{document}
\title{CNNs, Transformers, Hybrid, and Vision Language Models for Skin Cancer Detection\thanks{Accepted at ICPRAI 2026, The Fifth International Conference on Pattern Recognition and Artificial Intelligence. To appear in Lecture Notes in Computer Science.}}
\titlerunning{Comparing CNNs, Transformers, Hybrids, and VLMs}
%

\author{Durjoy Dey\,\orcidlink{0009-0002-9026-7202}\inst{1,2} \and Yuhong Yan\,\orcidlink{0009-0004-5572-1187}\inst{1} \and Hassan Hajjdiab\,\orcidlink{0000-0002-5661-8606}\inst{1}}
\authorrunning{D. Dey et al.}

\institute{
Department of Computer Science and Software Engineering, Concordia University, Montreal, Canada\\
\email{durjoy.dey@mail.concordia.ca}\\
\email{yuhong.yan@concordia.ca}\\
\email{hassan.hajjdiab@concordia.ca}
\and
Ebovir Biotechnologie Inc., Montreal, Canada
}

\maketitle              
\begin{abstract}
Skin cancer is a common and fast rising malignancy worldwide. Early detection is critical for improving outcomes. Deep learning models trained on dermoscopic and clinical images can support automated and fast triage. However, many studies evaluate only a limited set of architectures. Experimental setups also vary across studies. In this paper, we present a unified evaluation of twelve deep learning models for binary skin cancer detection on the PAD-UFES-20 dataset. The models span four families: convolutional neural networks (CNN), vision transformers (ViT), hybrid convolution transformer backbones, and vision language models (VLM).

Performance is assessed using AUC, the maximum F1 score with its precision and recall, and sensitivity at 80\% specificity, reflecting screening oriented requirements. Our results show that well tuned CNNs already provide strong baselines, but transformer based families consistently improve discrimination. Hybrid models (MaxViT Tiny, CoAtNet0) and a SigLIP based VLM achieve the best overall trade off between ranking performance and clinically relevant operating points, while CLIP based model offers high precision. The full codebase for all experiments is publicly released. Together, these findings offer practical guidance on which model families are most suitable for real world deployment in skin cancer screening and establish a reproducible reference point for future work on PAD-UFES-20.

\end{abstract}

\section{Introduction}
Skin is the largest organ of the human body \cite{1} and is constantly exposed to environmental insults, contributing to skin cancer being one of the most common malignancies worldwide \cite{2}. Its incidence is rising rapidly and it can be aggressive and potentially fatal, with substantial public health impact \cite{5}. Timely detection improves treatment success and helps prevent metastatic spread \cite{6}. In practice, clinicians rely on visual inspection and confirm suspected malignancy by biopsy \cite{7}. However, early detection remains difficult because lesion appearances vary widely and subtle lesions can be missed, even by experts \cite{9}. These limitations motivate more reliable detection approaches.

Deep learning now supports automated dermatological assessment, and CNN and transformer models show strong performance for skin cancer detection \cite{10}. Earlier hand crafted feature methods were sensitive to image quality and design choices \cite{21}. CNNs enable end to end learning from pixels but can miss long range context \cite{23}. Vision Transformers replace convolution with self attention over image patches and can match or outperform CNNs in dermatology by modelling long range lesion context \cite{24,25,26}. Hybrid backbones combine convolutional stems with attention blocks to retain CNN inductive bias while enabling global reasoning. Vision language models such as CLIP and SigLIP use large scale image text pretraining to learn transferable features that can be adapted to medical imaging with limited labels \cite{11}.

Evidence remains limited on how these architectural families compare under a unified, clinically oriented evaluation protocol. Most existing studies focus on CNNs or specific ViT variants, while hybrid convolution transformer backbones and vision language models are rarely evaluated side by side on realistic primary care datasets. Our goal is twofold: to quantify the performance gains of attention based and vision language models over strong CNN baselines under identical training and evaluation conditions, and to identify which families provide the best trade off between discrimination and screening relevant operating points. We therefore compare CNNs, vision transformers, hybrid models, and vision language models on PAD-UFES-20 for binary cancer versus non cancer classification. This analysis clarifies which architectural choices are most promising for reliable early detection tools in routine practice.

\section{Related Work}
Automated skin lesion analysis has moved from hand crafted features with classical machine learning to end to end deep learning. Early pipelines combined segmentation with color and texture descriptors and classifiers such as support vector machines or random forests, but remained sensitive to image quality and feature design \cite{Hardie2018HandCrafted}. Deep CNNs trained from pixels have reached dermatologist level performance and enabled clinical decision support \cite{Esteva2017,Brinker2019}. For example, Esteva et al. trained Inception v3 on over 129,000 clinical images and reported ROC performance for keratinocyte carcinoma and melanoma comparable to board certified dermatologists \cite{Esteva2017}. Later studies confirmed strong results with residual and Inception style networks using ImageNet pretraining, augmentation, and fine tuning \cite{Brinker2019,Tschandl2020}. However, CNNs are inherently local and may miss long range context.

Vision Transformers (ViTs) are an alternative for skin lesion classification and replace convolution with self-attention over image patches. Recent studies report that ViT-based models achieve strong performance for skin cancer segmentation and classification when trained and regularized appropriately \cite{ViTSkin2024}. Pacal et al. used a Swin-Transformer architecture with shifted-window self-attention and a SwiGLU MLP and reported 89.36\% accuracy and 86.61\% F1 on the eight-class ISIC 2019 dataset, outperforming many CNN and transformer baselines \cite{Pacal2024SwinSkin}. Self-attention helps model relationships between distant regions, supporting asymmetric patterns, irregular borders, and lesion context.

Beyond pure CNN and ViT architectures, hybrid convolution transformer backbones aim to combine CNN local inductive bias with global self attention and perform well on large vision benchmarks. However, most dermatology studies still compare only CNNs and ViTs. Ye et al. evaluated multiple CNNs, several ViT variants, and a CLIP based VLM on ISIC and reported that transformers and VLMs can outperform CNNs in accuracy and robustness \cite{YeICIIT2025}. Their study did not include hybrid architectures and focused on multiclass ISIC images rather than binary screening with smartphone photographs. We extend this line of work by evaluating hybrids together with CNNs, ViTs, and VLMs on PAD-UFES-20 under one framework, and we show that hybrids can match or exceed pure transformers and VLMs at clinically oriented operating points.

Vision language models (VLMs) have become a strong paradigm in medical image analysis. Models such as CLIP combine a high capacity vision encoder with a text encoder trained contrastively on large image caption corpora and can be adapted to medical tasks with lightweight fine tuning \cite{MedCLIP2022}. Wu et al. benchmarked CLIP ViT B/16 on MedMNIST and reported strong results, including an AUC of 0.9831 on DermaMNIST under end to end fine tuning \cite{Wu2025MedMNISTFoundation}. These findings suggest VLM vision encoders can match or surpass supervised CNN baselines with less task specific annotation. However, dermatology applications remain limited, and systematic evaluations on realistic datasets, such as PAD-UFES-20, are still scarce.

Prior work on PAD-UFES-20 mainly evaluates conventional CNN baselines or classical pipelines that combine image features with metadata \cite{PADUFES2020}. Hybrid convolution transformer backbones and vision language models are rarely assessed, and comprehensive side by side comparisons of CNNs, ViTs, hybrids, and VLMs under a unified protocol are lacking. Earlier studies also focus on generic metrics such as accuracy, while screening oriented measures like sensitivity at fixed specificity receive less attention. We address this gap with a unified pipeline evaluating twelve architectures across four families: CNNs, pure ViTs, hybrid convolution transformer backbones, and VLMs. All models are trained and tested on the same patient level splits of PAD-UFES-20 \cite{PADUFES2020} using consistent augmentation, loss functions, and early stopping. We compare models using AUC, F1$_\text{max}$ with its associated precision and recall, and sensitivity at 80\% specificity. This setup provides a controlled assessment of how these architectural choices affect cancer versus non cancer discrimination in a realistic primary care dataset.

\section{Methodology}
\begin{figure}[t]
    \centering
    \includegraphics[width=\textwidth]{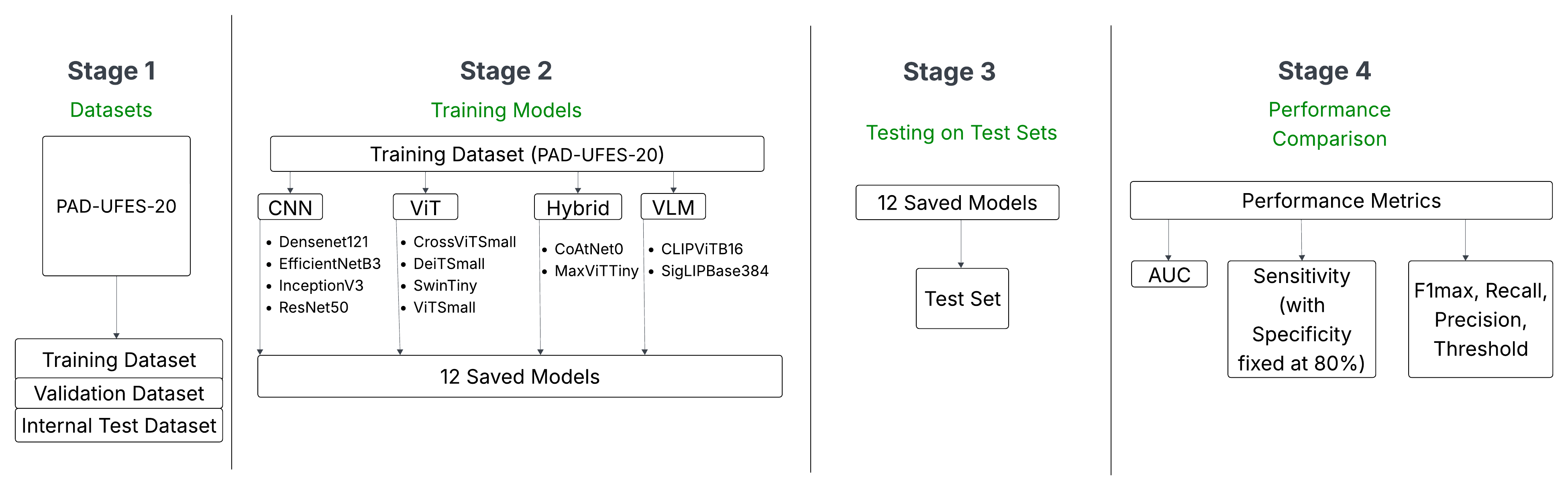}
    \caption{Experimental pipeline. (1) PAD-UFES-20 is split into patient-level train, validation, and test sets. (2) Twelve models from four families (CNN, ViT, hybrid, VLM) are trained on the training set. (3) The best checkpoint is evaluated on the test set. (4) Models are compared using AUC, sensitivity at 80\% specificity, and F1$_\text{max}$ with the associated precision, recall, and threshold.}
    \label{fig:method_pipeline}
\end{figure}

In this section, we describe the pipeline used to compare twelve deep learning architectures on PAD-UFES-20 for binary cancer versus non cancer detection. As shown in Fig.~\ref{fig:method_pipeline}, images are split at the patient level into training, validation, and test sets to prevent leakage. Images are then preprocessed and augmented using model specific transformations aligned with ImageNet or CLIP. We evaluate four model families (CNN, ViT, hybrid, and VLM) by adapting their final layers for binary prediction. All models are trained with the same optimization protocol and tested with AUC, sensitivity at 80\% specificity, and F1$_\text{max}$.

\subsection{Dataset}

We use the public PAD-UFES-20 dataset, which includes dermoscopic and clinical smartphone images from Brazilian primary care with lesion metadata and clinical or histopathological diagnoses \cite{PADUFES2020}. It contains 1,373 patients, 1,641 lesions, and 2,298 \texttt{.png} images with varying resolutions due to different devices. The skin lesions are annotated as basal cell carcinoma (BCC), squamous cell carcinoma (SCC), actinic keratosis (ACK), seborrheic keratosis (SEK), melanoma (MEL), and melanocytic nevus (NEV), yielding six categories: cancers (BCC, MEL, SCC) and non malignant conditions (ACK, NEV, SEK) \cite{PADUFES2020}.

We frame the task as binary cancer versus non cancer classification. Lesions labeled BCC, MEL, or SCC are assigned \texttt{is\_cancer=1}, and ACK, NEV, or SEK are assigned \texttt{is\_cancer=0}. All models use this binary label, with the original categories used only to define the mapping. The metadata includes lesion attributes and unique patient and lesion identifiers. We create a patient level split with 70\% train, 15\% validation, and 15\% test, ensuring no patient appears in multiple sets. Metrics are reported on the held out test set at the lesion level.

\subsection{Data Preprocessing}

We load raw RGB images from the PAD-UFES-20 folders as three channel tensors. We do not apply cropping, segmentation, or artifact removal, so models operate directly on routine clinical and dermoscopic photographs. Resizing and normalization follow each model family. For CNNs and transformers from \texttt{timm}, we use the official pretrained data configuration via \texttt{resolve\_data\_config} and \texttt{create\_transform}. For ViTs we apply stronger training augmentation, while validation and test use only resizing, center cropping, and ImageNet normalization. For CLIP models, training uses OpenCLIP preprocessing with random resized crops (224 or 336), horizontal flips, mild color jitter, and CLIP normalization, while evaluation uses resize, center crop, and the same normalization. For SigLIP, we use the official \texttt{timm} data configuration for the backbone. All models process floating point tensors normalized with their respective statistics.

\subsection{Model Selection and Architecture}

We compare four families of architectures: classical convolutional neural networks, pure vision transformers, hybrid convolution–transformer backbones and vision language models. Our CNN baselines are ResNet50 \cite{7780459}, InceptionV3 \cite{inproceedings}, DenseNet121 \cite{8099726}, and EfficientNetB3 \cite{pmlr-v97-tan19a}, all ImageNet pretrained. For these models and for all transformer and hybrid backbones, we replace the original classifier with a shared binary head: dropout, a 256 unit fully connected layer with ReLU, dropout, and a final linear layer producing one logit. Using the same head enables a controlled comparison of the feature extractors.

The transformer models are SwinTiny \cite{swin}, ViTSmall16 \cite{vitSmall}, DeiTSmall16 \cite{DeitSmall}, and CrossViTSmall \cite{crossViTSmall}. We load all models from \texttt{timm} with ImageNet pretrained weights and use a drop path rate of 0.3. Each model uses the same binary classification head described above. Hybrid models include CoAtNet0 \cite{coatnet0} and MaxViTTiny \cite{MaxViT}, which use convolutional stems with attention blocks. We load both from \texttt{timm} with ImageNet pretrained weights and apply the same binary head, so differences come from the backbone.

The vision language models are CLIP with a ViT B 16 image encoder \cite{Radford2021LearningTV} and SigLIP Base 384 \cite{siglip}. For CLIP, we use prompt based classification for \texttt{is\_cancer} by encoding malignant lesion prompts into a single cancer prototype and computing logits from cosine similarity to image embeddings, with a learnable logit scale and class bias. For SigLIP, we attach the shared binary head, train it with the backbone frozen, then unfreeze the last transformer blocks for fine tuning. All models perform binary classification and output a single logit for \texttt{is\_cancer}, representing the predicted malignancy probability.

\subsection{Model Training and Optimization}

All experiments use a fixed seed. We train with AdamW and a cosine schedule with warmup, using separate parameter groups with lower learning rates and higher weight decay for pretrained backbones than for the classification head. Class imbalance is handled with weighted \texttt{BCEWithLogitsLoss}. Training uses mixed precision (\texttt{torch.amp}) with gradient scaling and clipping. We use early stopping on validation AUC with a patience of 10 epochs. For vision language models, we apply two stage fine tuning by first training lightweight heads with frozen encoders, then unfreezing the last blocks with a lower backbone learning rate. We also use gradient accumulation to increase effective batch size under memory constraints.

\subsection{Overfitting Mitigation}

To reduce overfitting and improve generalisation, we apply strong training augmentation. Transformer models use random resized crops, horizontal flips, colour jitter, and random erasing, while vision language models follow their original pre training augmentations. These transformations increase appearance variation and reduce memorisation. At the model level, CNN, transformer, and hybrid backbones are regularised with dropout in the classification head and a non zero drop path rate for transformers. We also apply stronger weight decay to pretrained backbones than to the shallow heads, limiting updates to feature extractors while allowing the head to adapt. We use early stopping on validation AUC with a patience of 10 epochs. For vision language models, we apply two stage fine tuning by training lightweight heads with frozen encoders, then unfreezing the last blocks with a lower learning rate. All experiments use a strict patient level split to prevent leakage and inflated performance estimates.

\subsection{Accurate Evaluation Metrics}

We evaluate models on the held out test set using threshold free and threshold dependent metrics. The main global metric is AUC, which measures how well malignant lesions are ranked above non malignant lesions across all thresholds. To assess specific operating points, we compute precision, recall, and F1$\text{max}$. For each model we sweep thresholds in $[0,1]$, compute precision and recall, and derive
\begin{equation}
\mathrm{F1} = \frac{2 \cdot \mathrm{precision} \cdot \mathrm{recall}}{\mathrm{precision} + \mathrm{recall}} .
\end{equation}
We report F1$\text{max}$ and the associated precision, recall, and threshold. These results are used in Table~\ref{tab:all_model_results} and in precision recall curves.

We also evaluate a screening operating point. On the validation set, we choose the threshold that gives about 80\% specificity from the ROC curve, then apply it to the test set to report sensitivity at 80\% specificity. Confusion matrices and Fig.~\ref{fig:sensitivity80_comparison} use this threshold. For vision language models, we apply temperature scaling on validation logits before selecting thresholds to improve calibration. All models are compared using test AUC, F1$_\text{max}$, and sensitivity at 80\% specificity. These metrics capture both overall discrimination and clinically relevant operating points, enabling fair comparison of CNN, transformer, hybrid, and vision language models on PAD-UFES-20.

\subsection{Code Availability}
All code for data preprocessing, model training, and evaluation on PAD-UFES-20 is available at \url{https://github.com/Durjoy001/Skin-NeuralNET}.

\section{Results}
\begin{figure}[htbp]
    \centering
    \includegraphics[width=\textwidth]{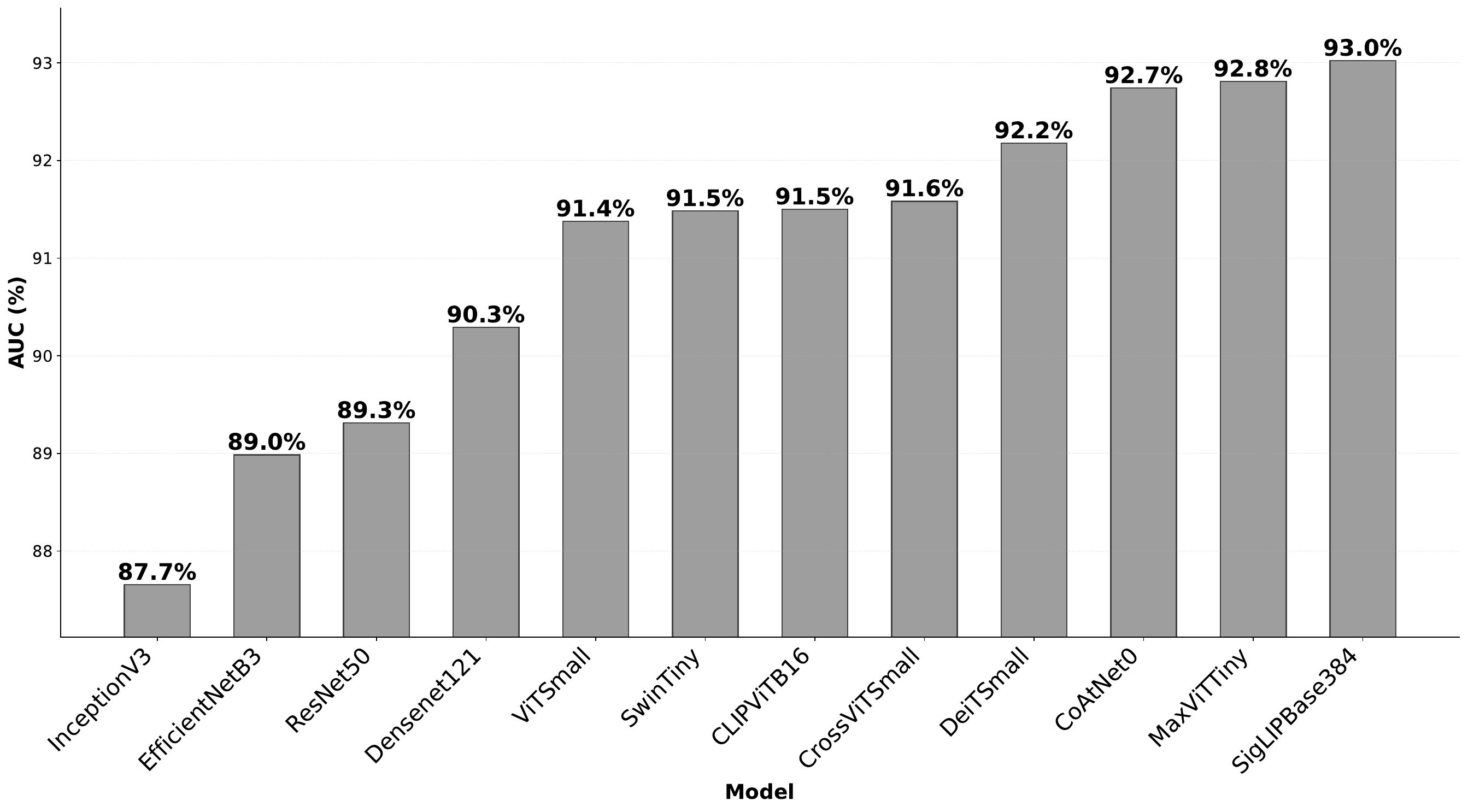}
    \caption{Test AUC comparison of CNN, ViT, hybrid, and vision language models on PAD-UFES-20. Higher values indicate better discrimination.}
    \label{fig:model_auc_comparison}
\end{figure}

\begin{figure}[t]
    \centering
    \includegraphics[width=\textwidth]{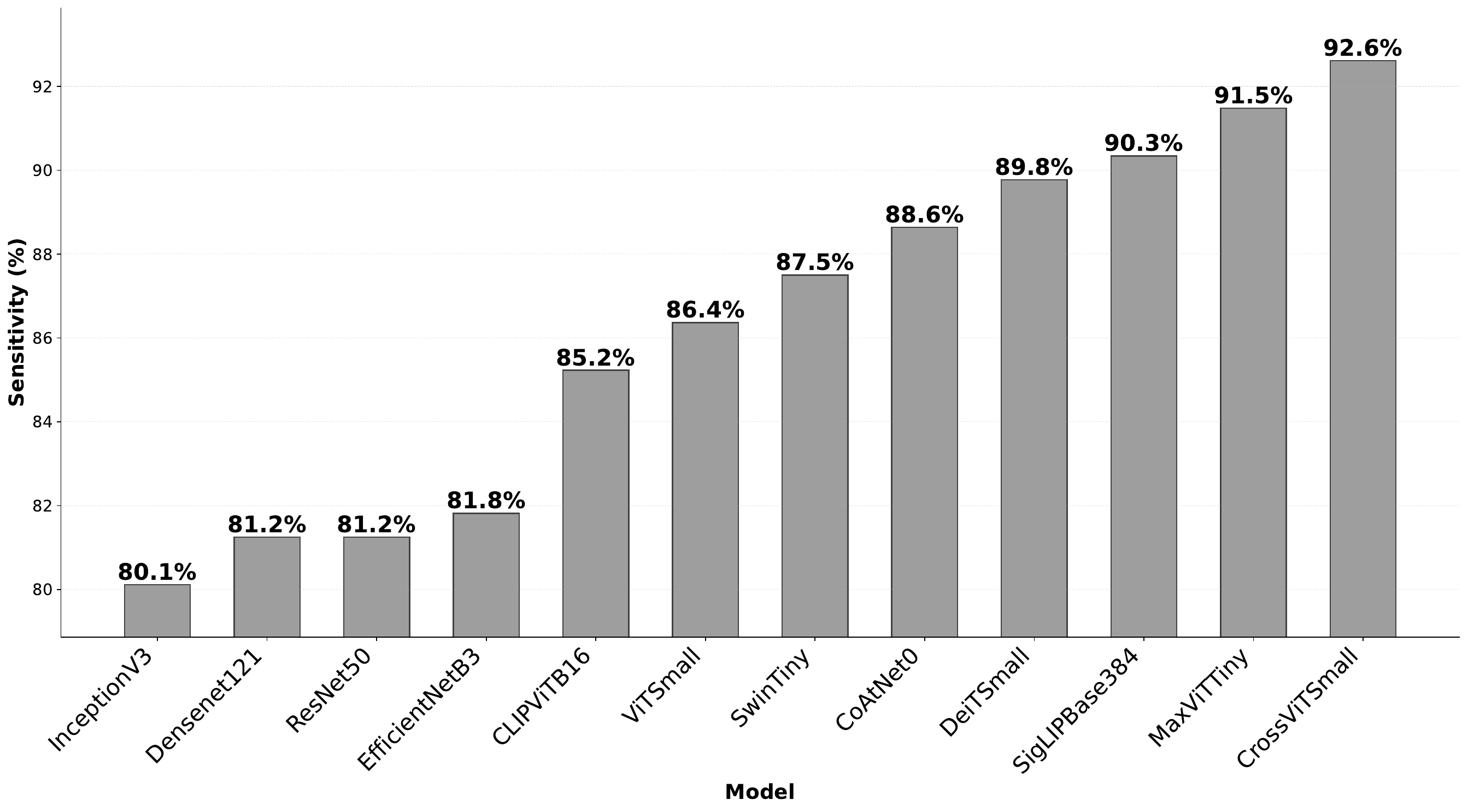}
    \caption{Model sensitivity at 80\% specificity on PAD-UFES-20.}
    \label{fig:sensitivity80_comparison}
\end{figure}

This section compares twelve models on PAD-UFES-20 for binary cancer versus non cancer classification across four families: CNNs, ViTs, hybrid backbones, and VLMs. We report test AUC and sensitivity at 80\% specificity (Figs.~\ref{fig:model_auc_comparison}, \ref{fig:sensitivity80_comparison}) and summarise results in Table~\ref{tab:all_model_results}. We then analyse each family and conclude with an overall comparison using AUC, F1$_\text{max}$ and sensitivity.

\begin{table}[t]
\caption{Test performance of all models on PAD-UFES-20 for cancer versus non cancer classification. Bold indicates the best value per metric.}
\label{tab:all_model_results}
\centering
\small
\renewcommand{\arraystretch}{1.1}%
\setlength{\tabcolsep}{6pt}

\resizebox{\textwidth}{!}{%
\begin{tabular}{|c|c|c|c|c|c|c|}
\hline
\textbf{Category} &
\textbf{Model} &
\textbf{AUC} &
\textbf{F1$_\text{max}$} &
\textbf{Precision} &
\textbf{Recall} &
\textbf{Threshold} \\
\hline
CNN   & Densenet121    & 0.902918 & 0.825974 & 0.760766 & 0.903409 & 0.262882 \\
CNN   & EfficientNetB3 & 0.889859 & 0.812500 & 0.812500 & 0.812500 & 0.541411 \\
CNN   & InceptionV3    & 0.876562 & 0.824147 & 0.765854 & 0.892045 & 0.313199 \\
CNN   & ResNet50       & 0.893131 & 0.816568 & 0.851852 & 0.784091 & 0.714025 \\
\hline
ViT   & CrossViTSmall  & 0.915829 & 0.867021 & 0.815000 & \textbf{0.926136} & 0.601837 \\
ViT   & DeiTSmall      & 0.921778 & 0.854749 & 0.840659 & 0.869318 & 0.617432 \\
ViT   & SwinTiny       & 0.914847 & 0.854054 & 0.814433 & 0.897727 & 0.327601 \\
ViT   & ViTSmall       & 0.913776 & 0.838889 & 0.820652 & 0.857955 & 0.524299 \\
\hline
Hybrid & CoAtNet0      & 0.927430 & 0.857143 & 0.845304 & 0.869318 & 0.655341 \\
Hybrid & MaxViTTiny    & 0.928085 & \textbf{0.869565} & 0.833333 & 0.909091 & 0.265923 \\
\hline
VLM   & CLIPViTB16     & 0.914996 & 0.839650 & \textbf{0.862275} & 0.818182 & \textbf{0.752202} \\
VLM   & SigLIPBase384  & \textbf{0.930227} & 0.863510 & 0.846995 & 0.880682 & 0.410952 \\
\hline
\end{tabular}%
}
\end{table}

\subsection{CNN Models Results Analysis}

The CNN baselines (DenseNet121, EfficientNetB3, InceptionV3, ResNet50) provide reference results for other model families on PAD-UFES-20. All are trained with the same patient level split for binary cancer versus non cancer classification. Their results are reported in Table~\ref{tab:all_model_results} and in Figs.~\ref{fig:model_auc_comparison} and \ref{fig:sensitivity80_comparison}.

DenseNet121 is the strongest CNN baseline, achieving the best AUC (0.903) and F1$\text{max}$ (0.826). ResNet50 and EfficientNetB3 follow in AUC (0.893 and 0.890), while InceptionV3 has the lowest AUC (0.877) but a similar F1$\text{max}$ (0.824). At a screening oriented operating point, CNNs reach competitive sensitivity at 80\% specificity (Fig.~\ref{fig:sensitivity80_comparison}), although the best transformer based and vision language models are slightly higher.

The CNNs also show distinct precision recall behaviour. Precision is the fraction of predicted cancers that are truly cancer, while recall is the fraction of true cancers that the model detects. DenseNet121 and InceptionV3 favour recall, achieving 0.903 and 0.892 recall with precision near 0.76, which supports screening scenarios where missing cancers is costly. ResNet50 is more conservative and prioritizes precision (0.852) at the expense of recall (0.784), which reduces false alarms but may miss more cancers. EfficientNetB3 provides a balanced trade off, with precision and recall both at 0.813. From a deployment perspective, these CNN backbones remain practical choices because they are generally lighter to run than large vision transformers or vision language encoders. This makes them attractive when compute or latency is constrained, while still providing strong screening performance on PAD-UFES-20.

\subsection{ViT Models Results Analysis}

The ViT models (CrossViTSmall, DeiTSmall, SwinTiny, ViTSmall) outperform the CNN baselines and narrow the gap to the hybrid and vision language architectures. Results are summarised in Table~\ref{tab:all_model_results} and in Figs.~\ref{fig:model_auc_comparison} and \ref{fig:sensitivity80_comparison}.

Vision Transformers provide a clear step up from CNN baselines on PAD-UFES-20. All ViT models achieve AUC above 0.91, led by DeiTSmall (0.922), with CrossViTSmall (0.916), SwinTiny (0.915), and ViTSmall (0.914) close behind. At a screening oriented operating point, ViTs also show higher sensitivity than CNNs at 80\% specificity (Fig.~\ref{fig:sensitivity80_comparison}), and they approach the strongest hybrid and vision language models.

Their precision recall behaviour is clinically meaningful. CrossViTSmall delivers the highest F1$_\text{max}$ among ViTs (0.867) and the highest recall (0.926) with precision 0.815, which supports settings where detecting as many cancers as possible is the priority. DeiTSmall and SwinTiny offer a more conservative trade off, with slightly lower recall but higher precision, reducing false alarms. ViTSmall is more balanced, with precision 0.821 and recall 0.858. In terms of deployment, ViTs typically require more computation than standard CNNs because self attention increases cost, especially at higher image resolutions. However, the consistent gains in AUC and sensitivity suggest that this extra compute can be justified when screening performance is the main goal.

\subsection{Hybrid Models Results Analysis}

Hybrid backbones provide the strongest results among the vision only models in this study. CoAtNet0 and MaxViTTiny both achieve AUC above 0.927, exceeding all CNN and ViT models, with MaxViTTiny slightly ahead (0.928085 vs 0.927430). At a screening oriented operating point, both hybrids also deliver among the highest sensitivities at 80\% specificity (Fig.~\ref{fig:sensitivity80_comparison}), meaning they detect more cancers while keeping the false positive rate fixed.

Their precision recall profiles are also strong. MaxViTTiny achieves the best F1$\text{max}$ of all models (0.869565), with precision 0.833333 and recall 0.909091, which supports screening settings that prioritise cancer detection. CoAtNet0 remains competitive (F1$\text{max}$ = 0.857143) and is slightly more precise (0.845304) with lower recall (0.869318), reflecting a more conservative balance. From a deployment perspective, these hybrid models are typically more computationally demanding than standard CNNs and often similar to, or heavier than, compact ViTs because they include both convolution and attention components. However, their consistent gains in AUC, F1$_\text{max}$, and sensitivity at 80

\subsection{VLM Models Results Analysis}

Vision language models also perform strongly when used as image only encoders on PAD-UFES-20. We evaluate CLIPViTB16 and SigLIPBase384 under the same binary classification setting, with results summarised in Table~\ref{tab:all_model_results} and compared in Figs.~\ref{fig:model_auc_comparison} and \ref{fig:sensitivity80_comparison}. SigLIPBase384 achieves the best AUC overall (0.930227) and a strong F1$\text{max}$ (0.863510), with precision 0.846995 and recall 0.880682. CLIPViTB16 is lower in AUC (0.914996) and F1$\text{max}$ (0.839650) but provides the highest precision of all models (0.862275) with recall 0.818182, reflecting a more conservative operating point.

At a screening oriented operating point, both VLMs achieve high sensitivity at 80\% specificity (Fig.~\ref{fig:sensitivity80_comparison}), with SigLIPBase384 combining strong discrimination with clinically relevant sensitivity. These results indicate that large scale vision language pretraining can yield image representations that are competitive with, and for SigLIPBase384 slightly better than, the best hybrid and ViT models even without using text at inference. In terms of deployment, VLM encoders are often heavier than CNNs and many compact ViTs, and they can require more memory and compute. However, their strong AUC and screening performance suggest they are a practical option when resources permit, and they offer a promising direction for improving automated skin lesion triage.

\subsection{Comparison of all models}

Table~\ref{tab:all_model_results} and Fig.~\ref{fig:model_auc_comparison} summarise results across all twelve models. AUC ranges from 0.876562 (InceptionV3) to 0.930227 (SigLIPBase384), and transformer based families (ViT, hybrid, VLM) consistently improve over CNNs by roughly two to four AUC points. For screening, the key outcome is how many cancers are detected while keeping false positives controlled. Hybrid and vision language models are strongest at this goal: SigLIPBase384 achieves the highest AUC (0.930227), and MaxViTTiny reaches the best F1$\text{max}$ (0.869565) with high recall (0.909091), indicating that it captures a large fraction of cancers at its optimal operating point. CoAtNet0 also performs near the top (AUC = 0.927430, F1$\text{max}$ = 0.857143).

Among ViTs, DeiTSmall and SwinTiny lead, and they improve sensitivity relative to CNN baselines, but remain slightly below the best hybrid and VLM models in overall screening performance. DenseNet121 is the strongest CNN baseline (AUC = 0.902918, F1$_\text{max}$ = 0.825974) and remains a practical option when compute is limited. In general, CNNs offer the lowest computational cost and easiest deployment, ViTs are typically heavier than CNNs, and hybrid and vision language encoders tend to be the most demanding. When resources permit and maximising cancer detection at fixed specificity is the priority, hybrid and vision language models provide the most favourable trade off. 

\section{Conclusions}

We evaluate twelve models on PAD-UFES-20 across CNN, ViT, hybrid, and VLM families for binary cancer versus non cancer classification on Brazilian primary care smartphone images. Using a shared patient level pipeline, we report AUC, F1$_\text{max}$, and sensitivity at 80\% specificity. These results suggest that models perform better when they can combine fine local lesion detail with broader global context, which likely explains the advantage of hybrid architectures and SigLIP based representations over standard CNN baselines. 

Our main contributions are threefold:
\begin{itemize}
    \item Unified benchmark. We provide, to the best of our knowledge, the first controlled study that evaluates CNNs, ViTs, hybrid backbones, and VLMs side by side on PAD-UFES-20 using the same patient-level splits, preprocessing, loss weighting, and optimisation settings.
    \item Evidence for hybrids and VLMs. CNNs provide strong baselines (AUC $\approx 0.90$), but hybrids and VLMs improve performance. MaxViTTiny achieves the best F1$\text{max}$, and SigLIPBase384 achieves the best AUC.
    \item Clinically oriented metrics and practical deployment. We report AUC, F1$_\text{max}$ with precision and recall, and sensitivity at 80\% specificity to reflect screening requirements. We also discuss computational trade offs across model families and release the full codebase to support reproducibility on PAD-UFES-20.
\end{itemize}

This study is limited to a single dataset with an internal patient level split, so external generalisability remains uncertain. The binary setting may mask subtype specific behaviour, and metadata, calibration, reliability, and efficiency analyses are left for future work. This comparison nevertheless provides practical guidance for deployment. When screening performance is the priority and resources allow, hybrid models such as MaxViTTiny and CoAtNet0 and the SigLIPBase384 encoder are strong defaults, with high AUC and favourable F1$_\text{max}$ and sensitivity at 80\% specificity. In resource constrained settings, compact CNNs such as DenseNet121 or ResNet50 remain competitive and simpler to deploy. Overall, this unified benchmark supports selecting model families that match clinical targets and operational constraints for real world screening.

\bibliographystyle{splncs04}
\bibliography{bibliography}

@article{Esteva2017,
  author  = {Esteva, Andre and Kuprel, Brett and Novoa, Roberto A. and others},
  title   = {Dermatologist-level classification of skin cancer with deep neural networks},
  journal = {Nature},
  year    = {2017},
  volume  = {542},
  number  = {7639},
  pages   = {115--118},
  doi     = {10.1038/nature21056}
}

@article{Tschandl2020,
  author  = {Tschandl, Philipp and Rinner, Christoph and Apalla, Zoe and others},
  title   = {Human--computer collaboration for skin cancer recognition},
  journal = {Nature Medicine},
  year    = {2020},
  volume  = {26},
  number  = {8},
  pages   = {1229--1234},
  doi     = {10.1038/s41591-020-0942-0}
}

@article{Brinker2019,
  author  = {Brinker, Titus J. and Hekler, Achim and Enk, Alexander H. and others},
  title   = {A convolutional neural network trained with dermoscopic images performs on par with 145 dermatologists in melanoma classification},
  journal = {European Journal of Cancer},
  year    = {2019},
  volume  = {111},
  pages   = {148--154},
  doi     = {10.1016/j.ejca.2019.01.011}
}

@article{ViTSkin2024,
  author  = {Galib Muhammad Shahriar Himel and Md. Masudul Islam and Kh Abdullah Al-Aff and others},
  title   = {Skin Cancer Segmentation and Classification Using Vision Transformer},
  journal = {Computational and Mathematical Methods in Medicine},
  year    = {2024},
  volume  = {2024},
  pages   = {3022192},
  doi     = {10.1155/2024/3022192}
}

@inproceedings{MedCLIP2022,
  author    = {Zhang, Jie and Liu, Xiaoxiao and Wang, Yuyin and others},
  title     = {MedCLIP: Contrastive Learning from Unpaired Images and Text for Medical Visual Representations},
  booktitle = {Proceedings of the IEEE/CVF Conference on Computer Vision and Pattern Recognition Workshops},
  year      = {2022}
}

@article{PADUFES2020,
  title={PAD-UFES-20: A skin lesion dataset composed of patient data and clinical images collected from smartphones},
  author={Andr{\'e} G. C. Pacheco and Gustavo Ribeiro Lima and Amanda da Silva Salom{\~a}o and others},
  journal={Data in Brief},
  year={2020},
  volume={32},
  doi={10.1016/j.dib.2020.106221}
}

@article{Hardie2018HandCrafted,
  title   = {Skin Lesion Segmentation and Classification for ISIC 2018 Using Traditional Classifiers with Hand-Crafted Features},
  author  = {Hardie, Russell C. and Ali, Redha and De Silva, Manawaduge Supun and others},
  journal = {arXiv preprint arXiv:1807.07001},
  year    = {2018}
}

@article{Pacal2024SwinSkin,
  author  = {Pacal, Ishak and Alaftekin, Melek and Zengul, Ferhat},
  title   = {Enhancing Skin Cancer Diagnosis Using Swin Transformer with Hybrid Shifted Window-Based Multi-Head Self-Attention and SwiGLU-Based MLP},
  journal = {Journal of Imaging Informatics in Medicine},
  year    = {2024},
  volume  = {37},
  pages   = {3174--3192},
  doi     = {10.1007/s10278-024-01140-8}
}

@inproceedings{YeICIIT2025,
  author    = {Ye, Chengwei and Li, Jiajun and Shuai, Qifan},
  title     = {Evaluating the Performance and Clinical Applications of Multiclass Deep Learning Models for Skin Cancer Pathology Diagnosis (ISIC): A Comparative Analysis of CNN, ViT, and VLM},
  booktitle = {Proceedings of the 2025 10th International Conference on Intelligent Information Technology (ICIIT 2025)},
  year      = {2025},
  pages     = {92--103},
  publisher = {Association for Computing Machinery},
  address   = {New York, NY, USA},
  doi       = {10.1145/3731763.3731793}
}

@article{Wu2025MedMNISTFoundation,
  author  = {Wu, Fuping and Papiez, Bart{\l}omiej W.},
  title   = {Rethinking Foundation Models for Medical Image Classification through a Benchmark Study on MedMNIST},
  journal = {arXiv preprint arXiv:2501.14685},
  year    = {2025},
  note    = {Under review for MIDL 2025}
}

@incollection{1,
  author    = {Walters, Kenneth A. and Roberts, Michael S.},
  title     = {The Structure and Function of Skin},
  booktitle = {Dermatological and Transdermal Formulations},
  editor    = {Walters, Kenneth A.},
  pages     = {19--58},
  publisher = {CRC Press},
  year      = {2002}
}

@article{2,
  author  = {Ferlay, Jacques and Colombet, Murielle and Soerjomataram, Isabelle and others},
  title   = {Cancer Statistics for the Year 2020: An Overview},
  journal = {International Journal of Cancer},
  year    = {2021},
  volume  = {149},
  number  = {4},
  pages   = {778--789}
}

@article{5,
  author  = {Fontanillas, Pierre and Alipanahi, Babak and Furlotte, Nicholas A. and others},
  title   = {Disease Risk Scores for Skin Cancers},
  journal = {Nature Communications},
  year    = {2021},
  volume  = {12},
  number  = {1},
  pages   = {160}
}

@article{6,
  author  = {Hussein, Mahmoud Rezk Abdelwahed},
  title   = {Skin Metastasis: A Pathologist's Perspective},
  journal = {Journal of Cutaneous Pathology},
  year    = {2010},
  volume  = {37},
  number  = {9},
  pages   = {e1--e20}
}

@article{7,
  author  = {Jerant, Anthony F. and Johnson, Jennifer T. and Sheridan, Catherine Demastes and others},
  title   = {Early Detection and Treatment of Skin Cancer},
  journal = {American Family Physician},
  year    = {2000},
  volume  = {62},
  number  = {2},
  pages   = {357--368}
}

@article{9,
  author  = {Kopf, Alfred W. and Salopek, Thomas G. and Slade, Johnny and others},
  title   = {Techniques of Cutaneous Examination for the Detection of Skin Cancer},
  journal = {Cancer},
  year    = {1995},
  volume  = {75},
  number  = {S2},
  pages   = {684--690}
}

@article{10,
  author  = {Furriel, Brunna C. R. S. and Oliveira, Bruno D. and Pro{\~a}, Renata and others},
  title   = {Artificial Intelligence for Skin Cancer Detection and Classification for Clinical Environment: A Systematic Review},
  journal = {Frontiers in Medicine},
  year    = {2024},
  volume  = {10},
  pages   = {1305954}
}

@inproceedings{11,
  author    = {Zhang, Pengchuan and Li, Xiujun and Hu, Xiaowei and Yang, Jianwei and others},
  title     = {VinVL: Revisiting Visual Representations in Vision--Language Models},
  booktitle = {Proceedings of the IEEE/CVF Conference on Computer Vision and Pattern Recognition (CVPR)},
  year      = {2021},
  pages     = {5579--5588}
}

@article{21,
  author  = {Anggriandi, D. and Utami, E. and Ariatmanto, D.},
  title   = {Comparative Analysis of {CNN} and {CNN-SVM} Methods for Classification Types of Human Skin Disease},
  journal = {Sinkron: Jurnal dan Penelitian Teknik Informatika},
  year    = {2023},
  volume  = {8},
  pages   = {2168--2178},
  doi     = {10.33395/sinkron.v8i4.12831}
}

@article{23,
  author  = {Agarwal, R. and Godavarthi, D.},
  title   = {Skin Disease Classification Using {CNN} Algorithms},
  journal = {EAI Endorsed Transactions on Pervasive Health and Technology},
  year    = {2023},
  volume  = {9},
  pages   = {1--8},
  doi     = {10.4108/eetpht.9.4039}
}

@article{24,
  author  = {Dhankar, U. and Jain, S. and Zaidi, S. and others},
  title   = {Skin Disease Detection Using Python and Deep Learning},
  journal = {International Journal of Engineering Applied Sciences and Technology},
  year    = {2023},
  volume  = {8},
  pages   = {186--191},
  doi     = {10.33564/ijeast.2023.v08i02.027}
}

@article{25,
  author  = {Aleissaee, A. A. and Kumar, A. and Anwer, R. M. and others},
  title   = {Transformers in Remote Sensing: A Survey},
  journal = {Remote Sensing},
  year    = {2023},
  volume  = {15},
  number  = {7},
  pages   = {1860},
  doi     = {10.3390/rs15071860}
}

@incollection{26,
  author    = {Lungu-Stan, V.-C. and Cercel, D.-C. and Pop, F.},
  title     = {{SkinDistilViT}: Lightweight Vision Transformer for Skin Lesion Classification},
  booktitle = {Advances in Intelligent Systems and Computing},
  series    = {Lecture Notes in Computer Science},
  volume    = {14254},
  publisher = {Springer Nature Switzerland},
  year      = {2023},
  doi       = {10.1007/978-3-031-44207-0_23}
}

@INPROCEEDINGS{7780459,
  author={He, Kaiming and Zhang, Xiangyu and Ren, Shaoqing and others},
  booktitle={2016 IEEE Conference on Computer Vision and Pattern Recognition (CVPR)}, 
  title={Deep Residual Learning for Image Recognition}, 
  year={2016},
  volume={},
  number={},
  pages={770-778},
  doi={10.1109/CVPR.2016.90}}

@inproceedings{inproceedings,
author = {Szegedy, Christian and Vanhoucke, Vincent and Ioffe, Sergey and others},
year = {2016},
month = {06},
pages = {},
title = {Rethinking the Inception Architecture for Computer Vision},
doi = {10.1109/CVPR.2016.308}
}

@INPROCEEDINGS{8099726,
  author={Huang, Gao and Liu, Zhuang and Van Der Maaten, Laurens and others},
  booktitle={2017 IEEE Conference on Computer Vision and Pattern Recognition (CVPR)}, 
  title={Densely Connected Convolutional Networks}, 
  year={2017},
  volume={},
  number={},
  pages={2261-2269},
  doi={10.1109/CVPR.2017.243}}

@InProceedings{pmlr-v97-tan19a,
  title = 	 {{E}fficient{N}et: Rethinking Model Scaling for Convolutional Neural Networks},
  author =       {Tan, Mingxing and Le, Quoc},
  booktitle = 	 {Proceedings of the 36th International Conference on Machine Learning},
  pages = 	 {6105--6114},
  year = 	 {2019},
  editor = 	 {Chaudhuri, Kamalika and Salakhutdinov, Ruslan},
  volume = 	 {97},
  series = 	 {Proceedings of Machine Learning Research},
  month = 	 {09--15 Jun},
  publisher =    {PMLR},
}

@INPROCEEDINGS{crossViTSmall,
  author={Chen, Chun-Fu Richard and Fan, Quanfu and Panda, Rameswar},
  booktitle={2021 IEEE/CVF International Conference on Computer Vision (ICCV)}, 
  title={CrossViT: Cross-Attention Multi-Scale Vision Transformer for Image Classification}, 
  year={2021},
  volume={},
  number={},
  pages={347-356},
  doi={10.1109/ICCV48922.2021.00041}}

@InProceedings{DeitSmall,
  title = 	 {Training data-efficient image transformers \& distillation through attention},
  author =       {Touvron, Hugo and Cord, Matthieu and Douze, Matthijs and others},
  booktitle = 	 {Proceedings of the 38th International Conference on Machine Learning},
  pages = 	 {10347--10357},
  year = 	 {2021},
  editor = 	 {Meila, Marina and Zhang, Tong},
  volume = 	 {139},
  series = 	 {Proceedings of Machine Learning Research},
  month = 	 {18--24 Jul},
  publisher =    {PMLR},
}

@INPROCEEDINGS{swin,
  author={Liu, Ze and Lin, Yutong and Cao, Yue and others},
  booktitle={2021 IEEE/CVF International Conference on Computer Vision (ICCV)}, 
  title={Swin Transformer: Hierarchical Vision Transformer using Shifted Windows}, 
  year={2021},
  volume={},
  number={},
  pages={9992-10002},
  doi={10.1109/ICCV48922.2021.00986}}

@ARTICLE{vitSmall,
  author={Lee, Seunghoon and Lee, Seunghyun and Song, Byung Cheol},
  journal={IEEE Access}, 
  title={Improving Vision Transformers to Learn Small-Size Dataset From Scratch}, 
  year={2022},
  volume={10},
  number={},
  pages={123212-123224},
  doi={10.1109/ACCESS.2022.3224044}}

@inproceedings{coatnet0,
author = {Dai, Zihang and Liu, Hanxiao and Le, Quoc V. and others},
title = {CoAtNet: marrying convolution and attention for all data sizes},
year = {2021},
isbn = {9781713845393},
publisher = {Curran Associates Inc.},
address = {Red Hook, NY, USA},
articleno = {303},
numpages = {13},
series = {NIPS '21}
}

@inproceedings{MaxViT,
  author    = {Tu, Zhengzhong and Talebi, Hossein and Zhang, Han and others},
  title     = {MaxViT: Multi-axis Vision Transformer},
  booktitle = {Computer Vision -- ECCV 2022},
  series    = {Lecture Notes in Computer Science},
  volume    = {13684},
  editor    = {Avidan, Shai and Brostow, Gabriel and Ciss{\'e}, Moustapha and Farinella, Giovanni Maria and Hassner, Tal},
  publisher = {Springer, Cham},
  year      = {2022},
  pages     = {459--479},
  doi       = {10.1007/978-3-031-20053-3_27}
}

@INPROCEEDINGS{siglip,
  author={Zhai, Xiaohua and Mustafa, Basil and Kolesnikov, Alexander and others},
  booktitle={2023 IEEE/CVF International Conference on Computer Vision (ICCV)}, 
  title={Sigmoid Loss for Language Image Pre-Training}, 
  year={2023},
  volume={},
  number={},
  pages={11941-11952},
  doi={10.1109/ICCV51070.2023.01100}}

@inproceedings{Radford2021LearningTV,
  title={Learning Transferable Visual Models From Natural Language Supervision},
  author={Alec Radford and Jong Wook Kim and Chris Hallacy and others},
  booktitle = {Proceedings of the 38th International Conference on Machine Learning},
  series    = {Proceedings of Machine Learning Research},
  volume    = {139},
  pages     = {8748--8763},
  year      = {2021},
  publisher = {PMLR}
}
\end{document}